\def\printing{normal}
\def\blind{blind}
\def\normal{normal}
\def\article{article}
\def\frontpage{frontpage}
\ifx\printing\blind\documentclass{ecai2008}\else
 \ifx\printing\normal\documentclass{ecai2008}\else
   \documentclass{article} \fi\fi
\usepackage{times}
\usepackage{epic}
\usepackage{ecltree}
\usepackage{graphicx}
\usepackage{latexsym}
 \usepackage{verbatim}

\begin{document}

\newlength{\halftextwidth}
\setlength{\halftextwidth}{0.47\textwidth}
\def\halffigsize{2.2in}
\def\thirdfigsize{1.5in}
\def\negvspace{0in}
\def\posvspace{0em}

\input epsf





\newcommand{\set}{\mathcal}
\newcommand{\myset}[1]{\ensuremath{\mathcal #1}}

\renewcommand{\theenumii}{\alph{enumii}}
\renewcommand{\theenumiii}{\roman{enumiii}}
\newcommand{\figref}[1]{Figure \ref{#1}}
\newcommand{\tref}[1]{Table \ref{#1}}
\newcommand{\And}{\wedge}

\newtheorem{mydefinition}{Definition}
\newtheorem{mytheorem}{Theorem}
\newtheorem{mytheorem1}{Theorem}
\newcommand{\myproof}{\noindent {\bf Proof:\ \ }}
\newcommand{\myqed}{\mbox{$\Box$}}

\newcommand{\mymod}{\mbox{\rm mod}}
\newcommand{\range}{\mbox{\sc Range}}
\newcommand{\roots}{\mbox{\sc Roots}}
\newcommand{\myiff}{\mbox{\rm iff}}
\newcommand{\alldifferent}{\mbox{\sc AllDifferent}}
\newcommand{\permutation}{\mbox{\sc Permutation}}
\newcommand{\disjoint}{\mbox{\sc Disjoint}}
\newcommand{\cardpath}{\mbox{\sc CardPath}}
\newcommand{\cardpathsec}{\mbox{CARDPATH}}
\newcommand{\CARDPATH}{\mbox{\sc CardPath}}
\newcommand{\common}{\mbox{\sc Common}}
\newcommand{\uses}{\mbox{\sc Uses}}
\newcommand{\lex}{\mbox{\sc Lex}}
\newcommand{\usedby}{\mbox{\sc UsedBy}}
\newcommand{\nvalue}{\mbox{\sc NValue}}
\newcommand{\slide}{\mbox{\sc Slide}}
\newcommand{\slidesec}{\mbox{SLIDE}}
\newcommand{\sliden}{\mbox{\sc Slide}}
\newcommand{\SLIDE}{\mbox{\sc CardPath}}
\newcommand{\circularslide}{\mbox{\sc CardPath}_{\rm O}}
\newcommand{\among}{\mbox{\sc Among}}
\newcommand{\mysum}{\mbox{\sc MySum}}
\newcommand{\amongseq}{\mbox{\sc AmongSeq}}
\newcommand{\atmost}{\mbox{\sc AtMost}}
\newcommand{\atleast}{\mbox{\sc AtLeast}}
\newcommand{\element}{\mbox{\sc Element}}
\newcommand{\gcc}{\mbox{\sc Gcc}}
\newcommand{\gsc}{\mbox{\sc Gsc}}
\newcommand{\contiguity}{\mbox{\sc Contiguity}}
\newcommand{\PRECEDENCE}{\mbox{\sc Precedence}}
\newcommand{\assignnvalues}{\mbox{\sc Assign\&NValues}}
\newcommand{\linksettobooleans}{\mbox{\sc LinkSet2Booleans}}
\newcommand{\domain}{\mbox{\sc Domain}}
\newcommand{\symalldiff}{\mbox{\sc SymAllDiff}}

\newcommand{\slidingsum}{\mbox{\sc SlidingSum}}
\newcommand{\MaxIndex}{\mbox{\sc MaxIndex}}
\newcommand{\REGULAR}{\mbox{\sc Regular}}
\newcommand{\regular}{\mbox{\sc Regular}}
\newcommand{\costregular}{\mbox{\sc CostRegular}}
\newcommand{\STRETCH}{\mbox{\sc Stretch}}
\newcommand{\SLIDEOR}{\mbox{\sc SlideOr}}
\newcommand{\NAE}{\mbox{\sc NotAllEqual}}
\newcommand{\myldots}{\mbox{..}}

\newcommand{\todo}[1]{{\tt (... #1 ...)}}

\title{\slidesec: A Useful Special Case \\ of the \cardpathsec\ Constraint}
\ifx\printing\blind
\author{
Tracking Number: 39\\}
\date{}
\else \fi

\ifx\printing\normal
\author{Christian Bessiere\institute{LIRMM (CNRS / U.  Montpellier), France,
email: bessiere@lirmm.fr. Supported 
by the ANR project ANR-06-BLAN-0383-02.} 
\and Emmanuel Hebrard\institute{4C, UCC,
Ireland, email: ehebrard@4c.ucc.ie.} \and Brahim
Hnich\institute{Izmir Uni. of Economics, Turkey, email:
brahim.hnich@ieu.edu.tr. Supported 
by the Scientific and Technological Research Council of Turkey (TUBITAK) 
under Grant No. SOBAG-108K027.} \and Zeynep Kiziltan\institute{CS
Department, Uni. of Bologna, Italy, email: zeynep@cs.unibo.it.} \and
Toby Walsh\institute{NICTA and UNSW, Sydney, Australia, email:
toby.walsh@nicta.com.au.
Funded by
the Australian Government's Department of Broadband, Communications and the 
Digital Economy,
and the ARC.}}
\else \fi
\ifx\printing\article
\author{Christian Bessiere\\
LIRMM \\
Montpellier, France \\
bessiere@lirmm.fr \And
Emmanuel Hebrard \\
4C\\
UCC, Ireland\\
ehebrard@4c.ucc.ie \And
Brahim Hnich\\
Izmir Uni. of\\
Economics, Turkey\\
brahim.hnich@ieu.edu.tr\And
Zeynep Kiziltan\\
CS Department \\
Uni. of Bologna, Italy \\
zeynep@cs.unibo.it \And
Toby Walsh\\
NICTA and UNSW\\
Sydney, Australia\\
tw@cse.unsw.edu.au}

\date{1st January 2007}
\else \fi

\ifx\printing\frontpage
\author{Christian Bessiere\\
LIRMM \\
Montpellier, France \\
bessiere@lirmm.fr \And
Emmanuel Hebrard \\
4C\\
UCC, Ireland\\
ehebrard@4c.ucc.ie \And
Brahim Hnich\\
Izmir Uni. of\\
Economics, Turkey\\
brahim.hnich@ieu.edu.tr\And
Zeynep Kiziltan\\
CS Department \\
Uni. of Bologna, Italy \\
zeynep@cs.unibo.it \And
Toby Walsh\\
NICTA and UNSW\\
Sydney, Australia\\
tw@cse.unsw.edu.au}
 \else \fi

\maketitle
\begin{abstract}
We study the \cardpath\ constraint. This ensures a given constraint
holds a number of times down a sequence of variables. We show that
\slide, a special case of \cardpath\ where the slid constraint must
hold always, can be used to encode a wide range of sliding sequence
constraints including \cardpath\ itself.
We consider how to propagate \slide\ and 
provide a complete
propagator for \cardpath . Since propagation is
NP-hard in general, we identify special cases where
propagation takes polynomial
time. 
Our experiments demonstrate that using \slide\ to encode 
global constraints can be as efficient and effective as
specialised propagators.
\end{abstract}

\ifx\printing\frontpage
\begin{center}
\begin{tabular}{ll}
Content areas: constraint programming, constraint satisfaction\\
\end{tabular}
\end{center}

\begin{quote}
This paper has not already been accepted by and is not currently under
review for a journal or another conference, nor will it be submitted
for such during ECAI's review period.
\end{quote}

\else \fi

\ifx\printing\frontpage
\eject \end{document} \else \fi

\section{INTRODUCTION}

In many scheduling problems, we have a
sequence of decision variables and a constraint which applies down
the sequence. For example, in the car sequencing problem, 
we need to decide the sequence of cars on a production line. We
might have a constraint on how often a particular option is met
(e.g. 1 out of 3 cars can have a sun-roof). 
As a second example, in a nurse rostering problem, we need
to decide the sequence of shifts worked by nurses. We might have a
constraint on how many consecutive night shifts any nurse can work.
Such constraints have been classified as sliding sequence
constraints \cite{beldiceanucatalog}.
To model such constraints, we can use the \cardpath\ constraint.
This ensures that a given constraint holds a number of times down a
sequence of variables \cite{cardinality-path}.
We identify a special case of \cardpath\ which we call \slide, 
that is interesting for several reasons. First,
many sliding sequence constraints, including \cardpath, can
easily be encoded using this special case.  
\slide\ is therefore a ``general-purpose"
constraint for encoding sliding sequencing constraints. This is
an especially easy way to provide
propagators for such global constraints
within a constraint toolkit. 
Second, we give a propagator
for enforcing generalised arc-consistency on \slide .
By comparison, the previous 
propagator for \cardpath\ given in \cite{cardinality-path}
does not prune all possible values. 
Third, 
\slide\ can be as efficient and effective as specialised propagators
in solving sequencing problems.

\section{\cardpathsec\ AND \slidesec\ CONSTRAINTS}

A constraint satisfaction problem consists of a set of variables,
each with a finite domain of values, and a set of constraints
specifying allowed combinations of values for given sets of 
variables. We
use capital letters for variables (e.g. $X$), and lower case for
values (e.g. $d$). We write $D(X)$ for the domain of 
variable $X$. Constraint solvers typically explore partial
assignments enforcing a local consistency property. A constraint
is \emph{generalised  arc  consistent} (\emph{GAC}) iff when a
variable is assigned any value in its domain, there exist
compatible values in the domains of all the other variables of the constraint. 

The \cardpath\ constraint was introduced in 
\cite{cardinality-path}. If $C$ is a constraint of arity $k$ then
$\cardpath(N,[X_1,\ldots,X_n],C)$ holds iff
$C(X_{i},\ldots,X_{i+k-1})$ holds $N$ times for $1 \leq i \leq
n-k+1$. For example, we can count the number
of changes in the type of shift with
$\cardpath(N,[X_1,\ldots,X_n],\neq)$. Note that \cardpath\ can be
used to encode a range of Boolean connectives since $N\geq1$ gives
disjunction, $N=1$ gives exclusive or, and $N=0$ gives negation.
We shall focus on a special case of the \cardpath\ constraint where
the slid constraint holds always.
$\slide(C,[X_1,\ldots,X_n])$ holds iff $C(X_{i},\ldots,X_{i+k-1})$
holds for all $1 \leq i \leq n-k+1$. That is, a \cardpath\
constraint in which $N=n-k+1$. We also
consider a more complex form of \slide\ that applies only every $j$
variables. More precisely, $\slide_j(C,[X_1,\ldots,X_n])$ holds iff
$C(X_{ij+1},\ldots,X_{ij+k})$ holds for $0 \leq i \leq
\frac{n-k}{j}$. By definition $\sliden_j$ for $j=1$ is equivalent to
$\sliden$.

Beldiceanu and Carlsson have shown  that \cardpath\ can 
encode a wide range of constraints like {\sc Change}, {\sc
Smooth}, {\sc AmongSeq} and {\sc SlidingSum}
\cite{cardinality-path}. As we discuss
later, \slide\ provides a simple way to encode such sliding sequencing
constraints. 
It can also encode many other more complex sliding 
sequencing constraints like \regular\ \cite{pesant1}, \STRETCH\
\cite{hpbcp2004}, and \lex\ \cite{beldiceanucatalog}, as well as
many types of chanelling constraints like \element\
\cite{hentenryck88} and optimisation constraints like the soft forms
of \regular\ \cite{vanHoeve1}. More interestingly, \cardpath\ can
itself be encoded into a \slide\ constraint.
In \cite{cardinality-path}, a propagator for \cardpath\ is
proposed that greedily
constructs upper and lower bounds on the number of (un)satisfied
constraints by posting and retracting (the negation of) each of
the constraints. This propagator does not achieve GAC. 
We propose here a complete propagator
for enforcing GAC on \slide . 
\slide\
thus provides 
a
GAC propagator for \cardpath. In addition,
\slide\ provides a GAC propagator for any of the other
global constraints it
can encode. As our experimental results reveal, \slide\ can be as
efficient and effective as specialised propagators.

We illustrate the usefulness of \slide\ with the \amongseq\
constraint 
which ensures that values occur with some given frequency. For
instance, we might want that no more than 3 out of every sequence of
7 shift variables are a ``night shift''. More precisely,
$\amongseq(l,u,k,[X_1,\ldots,X_n],v)$ holds iff between $l$ and
$u$ variables in every sequence of $k$ variables  take value in the
ground set $v$
\cite{beldiceanu5}. 
We can encode this using \sliden. More precisely,
$\amongseq(l,u,k,[X_1,\ldots,X_n],v)$ can be encoded as
$\slide(D^{k,v}_{l,u},[X_1,\ldots,X_n])$ where $D^{k,v}_{l,u}$ is an
instance of the \among\ constraint \cite{beldiceanu5}. 
$D^{k,v}_{l,u}(X_i,\ldots,X_{i+k-1})$ holds iff between $l$ and $u$
variables 
take values in the set $v$.
For example, 
suppose 2 of every 3 variables along a sequence $X_1 \ldots X_5$
should take the value $a$, where $X_1=a$ and $X_2,\ldots,X_5 \in
\{a,b\}$. This can be encoded as
$\sliden(E,[X_1,X_2,X_3,X_4,X_5])$ where $E(X_i,X_{i+1},X_{i+2})$ 
ensures two of its three
variables take $a$. This \slide\ constraint ensures that 
$E(X_1,X_2,X_3)$, $E(X_2,X_3,X_4)$ and
$E(X_3,X_4,X_5)$ all hold. Note that each ternary constraint is GAC.
However, enforcing GAC on the \sliden\ constraint sets $X_4=a$ as
there are only two satisfying assignments 
and neither have $X_4=b$. 

\section{\slidesec\ WITH MULTIPLE SEQUENCES}
\label{multipleseq}

We often wish to slide a constraint down two or more sequences of
variables at once. For example, suppose we want to ensure that two
vectors of variables, $X_1$ to $X_n$ and $Y_1$ to $Y_n$ differ at
{\em every} index. We can encode such a constraint by interleaving
the two sequences and sliding a constraint down the single sequence
with a suitable offset. In our example, we simply post
$\sliden_2(\neq,[X_1,Y_1,\ldots,X_n,Y_n])$.
As a second example of sliding down multiple sequences of variables,
consider the constraint $\REGULAR({\cal A},[X_1,\ldots,X_n])$. This
ensures that the values taken by a sequence of variables form a
string accepted by a deterministic finite automaton ${\cal A}$
\cite{pesant1}.
This global constraint is useful in scheduling, rostering and
sequencing problems to ensure certain patterns do (or do not) occur
over time. It can be used to encode a wide range of other global
constraints including: \among\ \cite{beldiceanu5}, \contiguity\
\cite{maher2002}, \lex\ 
and \PRECEDENCE\
\cite{llcp2004}.

To encode the \REGULAR\ constraint with \sliden, we introduce 
variables, $Q_i$ to record the state of the automaton. We
then post $\sliden_2(F,[Q_0,X_1,Q_1,\ldots,X_n,Q_n])$ where $Q_0$ is
set to the starting state, $Q_{n}$ is
restricted to accepting states, and
$F(Q_i,X_{i+1},Q_{i+1})$ holds iff $Q_{i+1}=\delta(X_{i},Q_i)$ where
$\delta$ is the transition function of the automaton. If we
decompose this encoding into the conjunction of slid constraints, we
get a set of constraints similar to 
\cite{Beldiceanu-Constraints05}. 
Enforcing GAC on this encoding ensures GAC on \regular\ and,
by exploiting functionaliy of $F$, takes
$O(ndq)$ time where $d$ is the number of values for $X_i$ and
$q$ is the number of states of the automaton. This is asymptotically
identical to the specialised \REGULAR\ propagator \cite{pesant1}.
This encoding is highly competitive 
in practice with the specialized propagator \cite{bhhkqwsara2007}. 

One advantage of this encoding 
is that it gives explicit access to the states of the automaton.
Consider, for example, a rostering problem where workers are allowed
to work for up to three consecutive shifts. 
This can be specified with a simple \REGULAR\ 
constraint. Suppose now we want to minimise the number of times a
worker has to work for three consecutive shifts. To encode this, we
can post an \among\ constraint on the state variables to count the
number of times we visit the state representing three consecutive
shifts, and minimise the value taken by this variable.
As we shall see later in the experiments, 
the encoding also gives an efficient {\em incremental} propagator.
In fact, the complexity of repeatedly
enforcing GAC on this encoding of the \REGULAR\
constraint down the whole branch of a backtracking search
tree is just $O(ndq)$ time.

\section{\slidesec\ WITH COUNTERS}
\label{counters}

We may want to slide a constraint on a sequence of
variables computing a count. We can use \slide\ to encode such
constraints by incrementally computing the count in an additional
sequence of variables. Consider, for example, 
$\cardpath(N,[X_1,\ldots,X_n],C)$. For simplicity, we consider
$k=2$
(i.e., $C$ is binary). The generalisation to other $k$
is straightforward. We introduce a sequence of integer variables
$M_i$ in which to accumulate the count. We encode \cardpath\ as
$\slide_2(G,[M_1,X_1,\dots,M_n,X_n])$ where $M_1=0$, $M_{n}=N$, and
$G(M_i,X_{i},M_{i+1},X_{i+1})$ is defined as: if $C(X_i,X_{i+1})$
holds then $M_{i+1}=M_i+1$, otherwise 
$M_{i+1}=M_i$. GAC on \slide\ ensures GAC on \cardpath.

As a second example, consider the \STRETCH\ constraint
\cite{hpbcp2004}. Given variables $X_1$ to
$X_n$ taking values from a set of shift types $\tau$, 
a set $\pi $ of ordered pairs
from $\tau \times \tau$, and functions $shortest(t)$
and $longest(t)$ giving the minimum and maximum length of
a stretch of type $t$,
$\STRETCH([X_1,\ldots,X_n])$ holds iff 
each stretch  of
type $t$ has length between $shortest(t)$ and
$longest(t)$; and  
consecutive types of stretches are in $\pi$. We can
encode \STRETCH\ as $\slide_2(H,[X_1, Q_1,\ldots,X_n,Q_n])$ where
$Q_1=1$ and $H(X_i,X_{i+1},Q_i,Q_{i+1})$ holds iff (1)
$X_i=X_{i+1}$, $Q_{i+1}=1+Q_i$, and $Q_{i+1} \leq longest(X_i)$; or
(2) $X_i \neq X_{i+1}$, $\langle X_i, X_{i+1} \rangle \in \pi$, $Q_i
\geq shortest(X_i)$ and $Q_{i+1}=1$. GAC on \slide\ ensures GAC on
\STRETCH.

\section{OTHER EXAMPLES OF \slidesec}\label{sec:others}


There are many other examples of global constraints which we can
encode using \sliden . 
For example, we can encode
\lex\ \cite{beldiceanucatalog} using \sliden. \lex\ holds iff a
vector of
variables $[X_1..
X_n]$ is lexicographically smaller than another
vector of variables $[Y_1..
Y_n]$. We  introduce a sequence of Boolean variables $B_i$ to
indicate if the vectors have been ordered by position $i-1$. Hence
$B_1=0$. We then encode \lex\ as
$\sliden_3(I,[B_1,X_1,Y_1,\ldots,B_n,X_n,Y_n])$ where
$I(B_i,X_i,Y_i,B_{i+1})$ holds iff $(B_i=B_{i+1}=0 \land X_i =
Y_i)$ or $(B_i=0\land B_{i+1}=1 \land X_i < Y_i)$ or
$(B_i=B_{i+1}=1)$.
This gives us a linear time propagator as efficient and incremental
as the specialised algorithm in \cite{fhkmwcp2002}. As a second
example, we can encode many types of channelling constraints using
\sliden\ like \domain\ \cite{refalo1}, \linksettobooleans\
\cite{beldiceanucatalog} and \element\ \cite{hentenryck88}. As a
final example, we can encode ``optimisation'' constraints like the
soft form of the \REGULAR\ constraint which measures the Hamming or
edit distance to a regular string \cite{vanHoeve1}.
There are, however, constraints that can be encoded using
\slide\ which do not give as efficient and effective propagators
as specialised algorithms (e.g. the global \alldifferent\ 
constraint \cite{regin1}).

\section{PROPAGATING \slidesec}
\label{sec:prop}

A constraint like \sliden\ is only really useful if we can
propagate it efficiently and effectively. 
The simplest possible way to propagate
$\sliden_j(C,[X_1,\ldots,X_n])$ is to decompose it into a
sequence of constraints, $C(X_{ij+1},\ldots,X_{ij+k})$ for $0 \leq
i \leq \frac{n-k}{j}$ and let the constraint solver propagate the
decomposition. Surprisingly,
this is enough to achieve GAC in many cases. 
For example, we can achieve GAC in this way on the \sliden\ encoding
of the 
\regular\ constraint.
If the constraints in the decomposition  overlap on just
one variable then the constraint graph 
is Berge acyclic   \cite{beeetal83}, and 
enforcing GAC on the decomposition of $\sliden_j$  achieves GAC on
$\sliden_j$.
Similarly, enforcing GAC on the decomposition achieves GAC on $\sliden_j$
if the constraint being slid is monotone. A constraint $C$ is
monotone iff there exists a total ordering $\prec$ of the 
values such that for any two values $v,w$, if $v \prec w$
then $v$ can replace $w$ 
in any support for $C$.
For instance, the constraints \among\ and {\sc Sum}
are monotone if either no upper bound, or no lower bound
is given.

\begin{mytheorem}
\label{gac-cond2} Enforcing GAC 
over each constraint in 
the decomposition of $\sliden_j$
achieves GAC on $\sliden_j$ if the constraint being slid is
monotone.
\end{mytheorem}
\myproof
For an arbitrary value $v \in D(X)$, we
show that if every constraint
is GAC, then we can build a support for $X=v$ on
$\sliden_j$. For any variable other than $X$,
we choose the smallest value in the total order.
This is the value that can be substituted for
any other value in the same domain. A tuple built this way
satisfies all the constraints being slid since we know that there
exists a support for each (they are GAC), and the values we chose can
be substituted for this support. \myqed

In the general case,
when
constraints overlap on more than one variable (e.g. in the \sliden\
encoding of \amongseq), we need to do more work to achieve GAC.
We distinguish two cases: when the arity of the constraint being
slid is not fixed, and when the arity is fixed. We show that
enforcing GAC in the former case is NP-hard.


\begin{mytheorem}
Enforcing GAC on $\sliden(C,[X_1,\ldots,X_n])$ is NP-hard when the
arity of $C$ is not fixed even if enforcing GAC on $C$ is itself
polynomial.
\end{mytheorem}
\myproof We give a reduction from 3-SAT in $N$ variables and $M$
clauses. We introduce variables $X^j_i$ for $1 \leq i
\leq N+1$ and $1 \leq j \leq M$. For each clause $j$, if the clause
is $x_a \vee \neg x_b \vee x_c$, then we set $X^j_1 \in \{x_a,
\neg x_b, x_c\}$ to represent the values that make this clause
true. For each clause $j$, we set $X^j_{i+1} \in \{0,1\}$ 
for $1 \leq i \leq N$ to represent a truth assignment. 
Hence, we duplicate the truth assignment for each
clause. We now build the following constraint
$\sliden(C,[X^1_1,\myldots,X^1_{N+1},
\myldots,X^j_1,\myldots,X^j_{N+1},\myldots, X^M_1,\myldots,X^M_{N+1}])$
where $C$ has arity $N+1$. We construct $C(Y_1,\ldots,Y_{N+1})$ 
to hold iff $Y_1 =
x_d$ and $Y_{1+d}=1$, or $Y_1 = \neg x_d$ and $Y_{1+d}=0$.
(in these two cases,
the value assigned to $Y_1$ represents the literal that
makes clause $j$ true), 
or $Y_i \in \{0,1\}$ and $Y_i=Y_{i+N+1}$ 
(in this case, the truth assignment is passed
down the sequence). 
Enforcing GAC on $C$ is polynomial and
an assignment 
satisfying the \sliden\ constraint
corresponds to a satisfying assignment for the original 3-SAT problem.
\myqed

When the arity of the constraint being slid is not great, we can
enforce GAC on \sliden\ using dynamic programming (DP) in a
similar way to the DP-based propagators for the \regular\ and
\STRETCH\ constraints \cite{pesant1,hpbcp2004}.
A much simpler method, however, which is just as efficient and
effective as dynamic programming is to exploit a variation of the
dual encoding into binary constraints 
\cite{dpaij89} based on tuples of support.
Such an encoding was proposed in \cite{bartak1}
for a particular sliding constraint. Here we show that this method
is more general and can be used for arbitrary \sliden\
constraints. Using such an encoding, \sliden\ can be easily added
to any constraint solver. We illustrate the intersection encoding
by means of an example.

\begin{figure}[tbp]
   \centering
   \includegraphics[width=8.5cm,height=3cm]{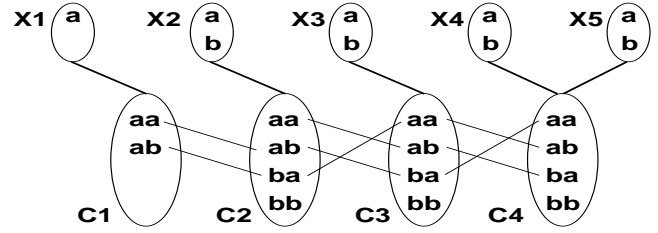} 
   \caption{Intersection encoding}
   \label{fig:intersect}
\end{figure}

Consider again the \amongseq\ example in which 2 of every 3
variables of $X_1 \ldots X_5$ should take the value $a$, where
$X_1=a$ and $X_2,\ldots,X_5 \in \{a,b\}$. We can encode this as
$\slide(E,[X_1,X_2,X_3,X_4,X_5])$ where $E(X_i,X_{i+1},X_{i+2})$ is
an instance of the \among\ constraint that ensures two of its three
variables take $a$.
If the sliding
constraint has arity $k$, we introduce an \emph{intersection}
variable for each subsequence of $k-1$ variables of \sliden. The
first intersection  variable $V_1$ has a domain containing all
tuples from $D(X_1)\times\ldots\times D(X_{k-1})$. The $j$th
intersection variable $V_j$ has domain containing $D(X_j)\times
\ldots\times D(X_{j+k-2})$. And so on until $V_{n-k+2}$. In our
example in Fig \ref{fig:intersect}, this gives $D(V_1)=D(X_1)\times
D(X_{2}),\ldots, D(V_4)= D(X_4)\times D(X_{5})$.
We then post binary compatibility constraints between consecutive
intersection variables. These  constraints ensure that the two
intersection variables assign $(k-1)$-tuples that agree on the
values of their $k-2$ common original variables (like constraints in
the dual encoding). They also ensure that the $k$-tuple formed
by the two $(k-1)$-tuples satisfies the corresponding instance of
the slid constraint. For instance, in Fig~\ref{fig:intersect},
the binary constraint between $V_1$ and $V_2$
does not allow the pair $\langle ab,aa\rangle$ because the second
argument of $ab$
for $V_1$ (value $b$ for $X_2$)
is in conflict with the first argument of $aa$ for $V_2$ (value $a$ for $X_2$).
That same constraint between $V_1$ and $V_2$
does not allow the pair  $\langle
ab,bb\rangle$ because  the tuple $abb$ is not allowed by $E(X_1,X_2,X_3)$.

Enforcing AC on such compatibility constraints prunes $aa$ and $bb$
from $V_2$, $ab$ and $bb$ from $V_3$, and $ba$ and $bb$ from $V_4$.
Finally, we post binary channelling constraints to link the tuples
to the original variables. One such constraint for each original
variable is sufficient. For example, we can have a channelling
constraint between $V_4$ and $X_4$ which ensures that the first
argument of the tuple assigned to $V_4$ equals the value assigned to
$X_4$. 
Enforcing AC on this channelling constraint prunes $b$ from the
domain of $X_4$. 
We could 
instead
post a channelling constraint between $V_3$
and $X_4$ ensuring that the second argument in $V_3$ equals $X_4$.
The \amongseq\ constraint is now GAC.
%

\begin{mytheorem}
Enforcing AC on the intersection encoding of $\sliden$ achieves GAC in
$O(nd^{k})$ time and $O(nd^{k-1})$ space where $k$ is the arity of the
constraint to slide and $d$ is the maximum domain size.
\end{mytheorem}
\myproof The constraint graph associated with the intersection
encoding is a tree. Enforcing AC on this therefore achieves GAC.
Enforcing AC on the channelling constraints then ensures that the
domains of the original variables are pruned appropriately. As we
introduce $O(n)$ intersection variables, and each can contain
$O(d^{k-1})$ tuples, the intersection encoding requires $O(nd^{k-1})$
space. Enforcing AC on a compatibility constraint between two
intersection variables $V_i$ and $V_{i+1}$ takes $O(d^k)$ time as each
tuple in the intersection variable $V_i$ has at most $d$ supports
which are the tuples of  $V_{i+1}$ that are equal to  $V_{i}$ on their
$k-2$ common arguments. Enforcing AC on $O(n)$ such constraints
therefore takes $O(nd^k)$ time. Finally, enforcing AC on each of the
$O(n)$ channelling constraints takes $O(d^{k-1})$ time as they are
functional. Hence, the total time complexity is $O(nd^k)$. \myqed\\

Arc consistency on the intersection encoding 
simulates
pairwise consistency on the decomposition. It does this efficiently 
as intersection variables represent in extension 'only' 
the intersections. This is sufficient because the constraint
graph is acyclic. This encoding is also 
very easy to implement in any constraint solver. It
 has good incremental properties. Only those constraints
associated with a variable which changes need to wake up.

The intersection encoding of $\slide_j$ for $j>1$ is less
expensive to build than for $j=1$
as we need intersection variables for
subsequences of  less than $k-1$ variables. For $1\leq j\leq k/2$,
we  introduce
intersection variables for subsequences of variables
of length $k-j$ starting at indices $1,j+1,2j+1...$ whose  domains contain
$(k-j)$-tuples of assignments. 
Compatibility
and  channelling
constraints are defined as with $j=1$.
If  $j>k/2$,   two consecutive intersection variables (for  two
subsequences of $k-j$ variables) involve less than $k$ variables of
the $\slide_j$. The compatibility constraint between them
cannot thus ensure the satisfaction of the slid constraint.
We therefore introduce intersection variables for
subsequences of length $\lceil k/2\rceil$
starting at indices $1,j+1,2j+1...$ and for
subsequences of length $\lceil k/2\rceil$
finishing at indices $k,j+k,2j+k...$
The compatibility constraint between two consecutive intersection
variables representing the subsequence starting at index $pj+1$ and
the subsequence finishing at index $pj+k$  ensures satisfaction of the
$(p+1)$th instance of the slid constraint.
The compatibility constraint between two consecutive intersection
variables representing subsequence finishing at index $pj+k$ and
the subsequence starting at index $(p+1)j+1$  ensures  the consistency
of the arguments in the intersection of two instances of the slid
constraint.

\section{EXPERIMENTS}



We now demonstrate the practical value of \slide. Due to 
space limits, 
we only report detailed results on a nurse scheduling problem,
and summarise the results on balanced incomplete block design
generation and car sequencing problems. Experiments are 
performed with
ILOG Solver 6.2 on a 2.8GHz Intel
computer running
Linux. 


We consider a Nurse Scheduling Problem \cite{burke}
in which we generate a schedule of shift
duties 
for a short-term planning period. There are
three types of shifts (day, evening, and night). We ensure that (1)
each nurse takes a day off or is assigned to an available
shift; (2) each shift has a minimum required number of nurses; (3)
each nurse's work load is between specific lower and upper
bounds; (4) each nurse works at most 5 consecutive days; (5) each
nurse has at least 12 hours of break between two shifts; (6)
the shift assigned to a nurse does not change more than once every
three days.
We construct four different models, all with variables
indicating what type of shift, if any, each 
nurse is working on each day. 
We break symmetry between the nurses with lex
concstraints. 
The constraints (1)-(3) are
enforced using global cardinality constraints. Constraints (4), (5)
and (6) form sequences of respectively 6-ary, binary and ternary
constraints. Since (4) is monotone, we simply post the
decomposition in the first three models. 
This achieves GAC by Theorem \ref{gac-cond2}. The models differ in
how (5) and (6) are propagated. In \texttt{decomp}, they are decomposed
into conjunction of slid constraints. In \texttt{amongseq}, (5) is
decomposed and (6) is enforced using the \amongseq\ constraint of
ILOG Solver (called \texttt{IloSequence}). The combination of
(5) and (6) are enforced by \slide\ in \texttt{slide}.
Finally, in
\texttt{slide$_c$}, we use \slide\ for the combination of (4), (5),
and (6).

We test the models using the instances available
at 
{\em http://www.projectmanagement.ugent.be/nsp.php} 
in which nurses have no maximum workload, but a set of preferences
to optimise. We ignore these preferences and post a  constraint
bounding the maximum workload to at most 5 day shifts, 4 evening
shifts and 2 night shifts per nurse and per week. Similarly, each
nurse must have at least 2 rest days per week. We solve three
samples of instances involving 25, 30 and 60 nurses to schedule over
28 days.

We use the same variable ordering for all models so that heuristic
choices do not affect results. We schedule the days in
chronological order and within each day we allocate a shift to every
nurse in lexicographical order. Initial experiments show that this
is more efficient than the minimum domain
heuristic. However, it restricts the variety of domains
passed to the propagators, and thus hinders
any demonstration of differences in pruning.
We therefore also use a more random heuristic.
We allocate within
each day a shift to every nurse randomly with $20\%$ frequency and 
lexicographically otherwise. 

\begin{table}
  \begin{scriptsize}
    \begin{center}
      \begin{tabular}{lrrrrr}
        \hline
        & \#solved & bts$^1$ & time$^1$ & bts$^2$ & time$^2$ \\
        &\multicolumn{5}{c}{25 nurses, 28 days (99 instances)}\\
        \hline
        \texttt{decomp} &    99 &        301 &     0.13 &        301 &     0.13\\
        \texttt{amongseq}  &    99 &        301 &     0.19 &        301 &     0.19\\
        \texttt{slide} &    99 &        301 &     0.19 &        301 &     0.19\\
        \texttt{slide$_c$} &    99 &        295 &     0.68 &        295 &     0.68\\
        \hline
        &\multicolumn{5}{c}{30 nurses, 28 days (99 instances)}\\
        \hline
        \texttt{decomp} &    68 &       7101 &      2.80 &      15185 &      5.29\\
        \texttt{amongseq} &    67 &       7101 &       4.31 &       7150 &      4.33\\
        \texttt{slide} &    70 &       3303 &      1.99 &       4319 &       2.53\\
        \texttt{slide$_c$} &    75 &       1047 &      2.13 &      11014 &      10.02\\
        \hline
        &\multicolumn{5}{c}{60 nurses, 28 days (100 instances)}\\
        \hline
        \texttt{decomp} &    51 &       5999 &      4.38 &       5999 &      4.38\\
        \texttt{amongseq} &    51 &       5999 &      7.10 &       5999 &      7.10\\
        \texttt{slide} &    52 &       5300 &      5.61 &       8479 &       7.21\\
        \texttt{slide$_c$} &    58 &       2157 &      7.52 &       4501 &      12.07\\
        \hline
      \end{tabular}
    \end{center}
    \vspace{-.7cm}
    \caption{\label{table-NSPlex}Nurse scheduling with lexicographical variable ordering 
    ($^1$ on instances solved by all methods, $^2$ on instances solved by the method).}
  \end{scriptsize}
\end{table}
\begin{table}
  \begin{scriptsize}
    \begin{center}
      \begin{tabular}{lrrrrr}
        \hline
        & \#solved & bts$^1$ & time$^1$ & bts$^2$ & time$^2$ \\
        &\multicolumn{5}{c}{25 nurses, 28 days (99 instances)}\\
        \hline
        \texttt{decomp} &    86 &      35084 &      7.69 &      41892 &      10.06\\
        \texttt{amongseq} &    85 &      35401 &      14.43 &      35401 &      14.43\\
        \texttt{slide} &    97 &       1699 &      1.00 &       1547 &     0.92\\
        \texttt{slide$_c$} &    97 &        457 &     0.58 &        438 &     0.56\\
        \hline
        &\multicolumn{5}{c}{30 nurses, 28 days (99 instances)}\\
        \hline
       \texttt{decomp} &    20 &      68834 &      11.94 &      69550 &      12.75\\
        \texttt{amongseq} &    20 &      68834 &      18.89 &      69550 &      19.83\\
        \texttt{slide} &    42 &        378 &     0.18 &       8770 &      7.29\\
        \texttt{slide$_c$} &    43 &        365 &     0.95 &      12857 &      6.76\\
        \hline
        &\multicolumn{5}{c}{60 nurses, 28 days (100 instances)}\\
        \hline
        \texttt{decomp} &     3 &     122406 &      71.06 &     250427 &      142.90\\
        \texttt{amongseq} &     2 &     122406 &      119.40 &     122406 &      119.40\\
        \texttt{slide} &    27 &        562 &      0.65 &       2367 &      2.19\\
        \texttt{slide$_c$} &    34 &        542 &       3.96 &       1368 &      6.38\\
        \hline
      \end{tabular}
    \end{center}
    \vspace{-.7cm}
    \caption{\label{table-NSPrandom}Nurse scheduling with random variable ordering 
    ($^1$ on instances solved by all methods, $^2$ on instances solved by the method).}
  \end{scriptsize}
\end{table}

Tables~\ref{table-NSPlex} and \ref{table-NSPrandom} report the mean
runtime and fails to solve the instances with 5 minutes
cutoff. Between the
first three models, the best results are due to \texttt{slide}. 
We solve more instances with \texttt{slide}, 
as well as explore a smaller tree. 
By developing a propagator for a 
generic constraint like \slide, we can 
increase pruning without hurting efficiency. Note that \texttt{slide}
always
performs better than \texttt{amongseq}. A possible reason is that
\amongseq\ cannot encode constraint (6) as directly as \slide. 
As in previous models, we need to channel 
into Boolean variables and post \amongseq\ on them. This may not
give as effective and efficient pruning. 
\slide\ thus offers both modelling and solving advantages over
existing sequencing constraints. Note also that \texttt{slide$_c$}
solves additional instances in the time limit. This is not suprising as the
model slides the combination of the constraints (4), (5), and (6).
Recall that the sliding constraint of (4) is 6-ary. 
It is pleasing to note that the intersection encoding performs well
even in the presence of such a high arity constraint. 

We also ran experiments on Balanced Incomplete Block Designs (BIBDs)
and car sequencing. 
For BIBD,
we use the model in \cite{fhkmwcp2002} which contains
\lex\ constraints. We propagate these either using the specialised
algorithm of \cite{fhkmwcp2002} or the \sliden\ encoding.
As both propagators maintain GAC, we only compare runtimes. 
Results on large instances show that the \slide\ model is as
efficient as the \lex\ model.
For car sequencing, we test
the scalability of \slide\ on large arity constraints and 
large domains using
$80$ instances from CSPLib. 
Unlike a model 
using \texttt{IloSequence}, our \slide\ model does not combine
reasoning about overall cardinality of a configuration
with the sequence of \among\ constraints. Hence, it 
is not as efficient:
$26$ instances were solved with \slide\ within the five 
minute cutoff, 
compared to $39$ with \texttt{IloSequence}. However, $9$ of the 
instances solved with \slide\ were not solved 
by \texttt{IloSequence}. 
The memory overhead of the \sliden\ propagator was not 
excessive despite
the slid constraints having arity $5$ and domains of 
size $30$.
The \slide\ model used on average $22$Mb of space, 
compared to $5$Mb for \texttt{IloSequence}.

\section{RELATED WORK}

Pesant introduced the \REGULAR\ constraint, and gave a propagator
based on dynamic programming to enforce GAC
\cite{pesant1}. As we saw, the \REGULAR\ constraint can be encoded
using a simple \slide\ constraint. 
In this simple case, the dynamic
programming machinery of Pesant's propagator
is unnecessary as the decomposition into
ternary constraints does not hinder propagation. 
We have found that \slide\ 
is as efficient as \regular\ in practice \cite{bhhkqwsara2007}.
Furthermore, our encoding introduces
variables for representing the states. Access to the state variables
may be useful (e.g. for expressing objective functions).
Although an objective function can be represented with the
\costregular\ constraint \cite{CostRegular}, this is limited to the
sum of the variable-value assignment costs. Our encoding is more
flexible, allowing different objective functions
like the min function used in the example in Section
\ref{multipleseq}.

Beldiceanu, Carlsson, Debruyne and Petit have proposed specifying
global constraints by means of deterministic finite automata
augmented with counters \cite{Beldiceanu-Constraints05}. They
automatically construct propagators for such automata by decomposing
the specification into a sequence of signature and transition
constraints. This gives an encoding similar to our \slide\
encoding of the \REGULAR\ constraint. There are, however, a number of
advantages of \slide\ over using an automaton. If the
automaton uses counters, pairwise consistency 
is needed to guarantee GAC (and most constraint 
toolkits do not support pairwise consistency). We can encode such
automata using a \slide\ where we introduce an additional sequence
of variables for each counter. \slide\ thus provides a GAC
propagator for such automata. Moreover, \slide\ has a better
complexity than a brute-force pairwise consistency algorithm based
on the dual encoding as it considers only the intersection variables,
reducing the space complexity by a factor of $d$.


Hellsten, Pesant and van Beek developed a GAC propagator for the
\STRETCH\ constraint based on dynamic programming
similar to that for the \REGULAR\ constraint \cite{hpbcp2004}. As we
have shown, we can encode the \STRETCH\ constraint and maintain GAC
using \slide.
Several propagators for the \amongseq\ are proposed and compared
in \cite{vanHoeve2,ascp07}. Among these propagators, those
based on the \regular\ constraint do the most
pruning and are often fastest. 
Finally,  Bartak
has proposed a similar intersection encoding
for propagating a sliding scheduling 
constraint \cite{bartak1}
We have shown that this method
is more general and can be used for arbitrary \sliden\
constraints. 

\section{CONCLUSIONS}

We have studied the \cardpath\ constraint. This slides a constraint
down a sequence of variables. We considered \slide\,
a special case of \cardpath\ in which the slid constraint holds at
every position. We demonstrated that this special case 
can encode many global sequencing constraints including 
\amongseq, \cardpath,
\regular\ in a simple way. \slide\ can
therefore serve as a ``general-purpose" constraint for decomposing a
wide range of global constraints, facilitating their integration
into constraint toolkits.
We proved that enforcing GAC on \slide\ is
NP-hard in general. Nevertheless, we identified
several useful and common cases where it is polynomial.
For instance, when the constraint 
being slid overlaps on just one variable or is monotone, 
decomposition does not hinder propagation. 
Dynamic programming or a variation of the dual encoding can be used
to propagate \slide\ when the constraint being slid overlaps on more
than one variable and is not monotone. 
Unlike the previous proposed propagator
for \cardpath, this achieves GAC. 
%
Our experiments demonstrated that using \slide\ to encode
constraints can be as efficient and effective as specialised
propagators. There are many directions for future work. One
promising direction is to use binary decision diagrams to store the
supports for the constraints being slid when they have many
satisfying tuples. We believe this could improve the efficiency of
our propagator in many cases.

\ifx\printing\article \vspace{-2mm}
\fi




\end{document}